\documentclass[letterpaper, 10 pt, conference]{ieeeconf}
\IEEEoverridecommandlockouts
\usepackage{cite}
\usepackage{bbold}
\usepackage{amsmath,amssymb,amsfonts}
\usepackage{booktabs}
\usepackage{algorithmic}
\usepackage{graphicx}
\usepackage{textcomp}
\usepackage{xcolor}
\usepackage{xspace}
\usepackage{adjustbox}
\usepackage{tabularx}
\def\BibTeX{{\rm B\kern-.05em{\sc i\kern-.025em b}\kern-.08em
    T\kern-.1667em\lower.7ex\hbox{E}\kern-.125emX}}

\newcommand{\softmax}{\ensuremath{\textrm{softmax}}}
\newcommand{\nuplan}{nuPlan\xspace}

\usepackage[absolute,showboxes]{textpos}

\begin{document}

\title{From Prediction to Planning With Goal Conditioned Lane Graph Traversals
}

\author{Marcel Hallgarten$^{1,2}$, Martin Stoll$^{1}$ and Andreas Zell$^{2}$
\thanks{$^{1}$Marcel Hallgarten and Martin Stoll are with Robert Bosch GmbH,
        Stuttgart, Germany
        {\tt\small marcel.hallgarten@de.bosch.com}
        {\tt\small martin.stoll@de.bosch.com}
        }%
\thanks{$^{2}$Marcel Hallgarten and Andreas Zell are with the Cognitive Systems Group, 
        University of T\"ubingen, T\"ubingen, Germany
         {\tt\small andreas.zell@uni-tuebingen.de}}%
}
\TPMargin*{3pt}
\newcommand\copyrighttext{
    \footnotesize
    \noindent
    \textcopyright\,2023 IEEE.
    Personal use of this material is permitted.
    Permission must be obtained for all other uses, in any current or future media, including reprinting/republishing this material for advertising or promotional purposes, creating new collective works, for resale or redistribution to servers or lists, or reuse of any copyrighted component of this work in other works.}%
\newcommand\copyrightnotice{%
    \begin{textblock*}{7in}(0.75in,0.15in)
        \copyrighttext
    \end{textblock*}
}
\maketitle
\copyrightnotice

\begin{abstract}
The field of motion prediction for automated driving has seen tremendous progress recently,
bearing ever-more mighty neural network architectures.
Leveraging these powerful models bears great potential for the closely related planning task.
In this work, we show that state-of-the-art prediction models can be converted into goal-directed planners. 
To this end, we propose a novel goal-conditioning method.
Our key insight is that conditioning prediction on a navigation goal at the behaviour level outperforms other widely adopted methods, with the additional benefit of increased model interpretability.
Moreover, our Method can be applied at inference time only. Hence, no ground-truth navigation command is required during training.
We evaluate our method on a large open-source dataset and show promising performance in a comprehensive benchmark.
Code is available under {\color{magenta} \texttt{https://mh0797.github.io/gc-pgp/}}.
\end{abstract}

\section{Introduction}
\label{introduction}
The enormous challenge of enabling autonomous driving is commonly subdivided into the tasks of perception, prediction, and planning.
While perception is supposed to determine the type and the location of objects and infrastructure based on raw sensor data such as camera images or LiDAR point clouds, prediction aims at forecasting the objects' future motion. Finally, a downstream planning algorithm's task is to navigate safely and comfortably towards a goal.

Learning-based Prediction and Planning algorithms exhibit large similarities regarding input and output representations.
Both aim to regress the trajectory that a target vehicle will follow based on a given scene context.
However, the fields differ in key aspects, namely whether the intention is assumed to be known and the requirements placed on the output trajectory(s).
In particular, in contrast to planning, prediction methods forecast the motion of observed agents solely based on current and past observations, \textit{i.e.,} without knowing their intention.
To account for the unknown intention, prediction methods commonly regress multiple trajectories that an agent could follow. 
Consequently, the diversity of the predicted trajectories, \textit{i.e.,} the number of captured potential intentions plays an important role.
In contrast, in the planning task, the controlled vehicle's intention is presumed to be known, making diversity at the intention level obsolete.
Moreover, as the controlled vehicle is assumed to follow the trajectory produced by a planning algorithm, collision avoidance, comfort and kinematic feasibility are fundamental requirements for a planner's output.

The former difference, namely the absence of intention in trajectory prediction, makes large datasets easy to collect.
Learning prediction models through supervised learning is straightforward.
Promoted by large open-source datasets with associated benchmarks~\cite{ettinger2021large, caesar2020nuscenes, wilson2023argoverse} the field of vehicle trajectory prediction has seen tremendous progress recently.
At the same time, learning a planner from these datasets is difficult, because observation alone without knowing the intention is not sufficient to unambiguously understand and imitate an expert's decision~\cite{codevilla2018end}.
Hence, carrying over advances from prediction to planning is a challenging task.

\begin{figure}
\centering
\includegraphics[width=0.8\linewidth, trim={0, 2.5cm 0, 0, clip}]{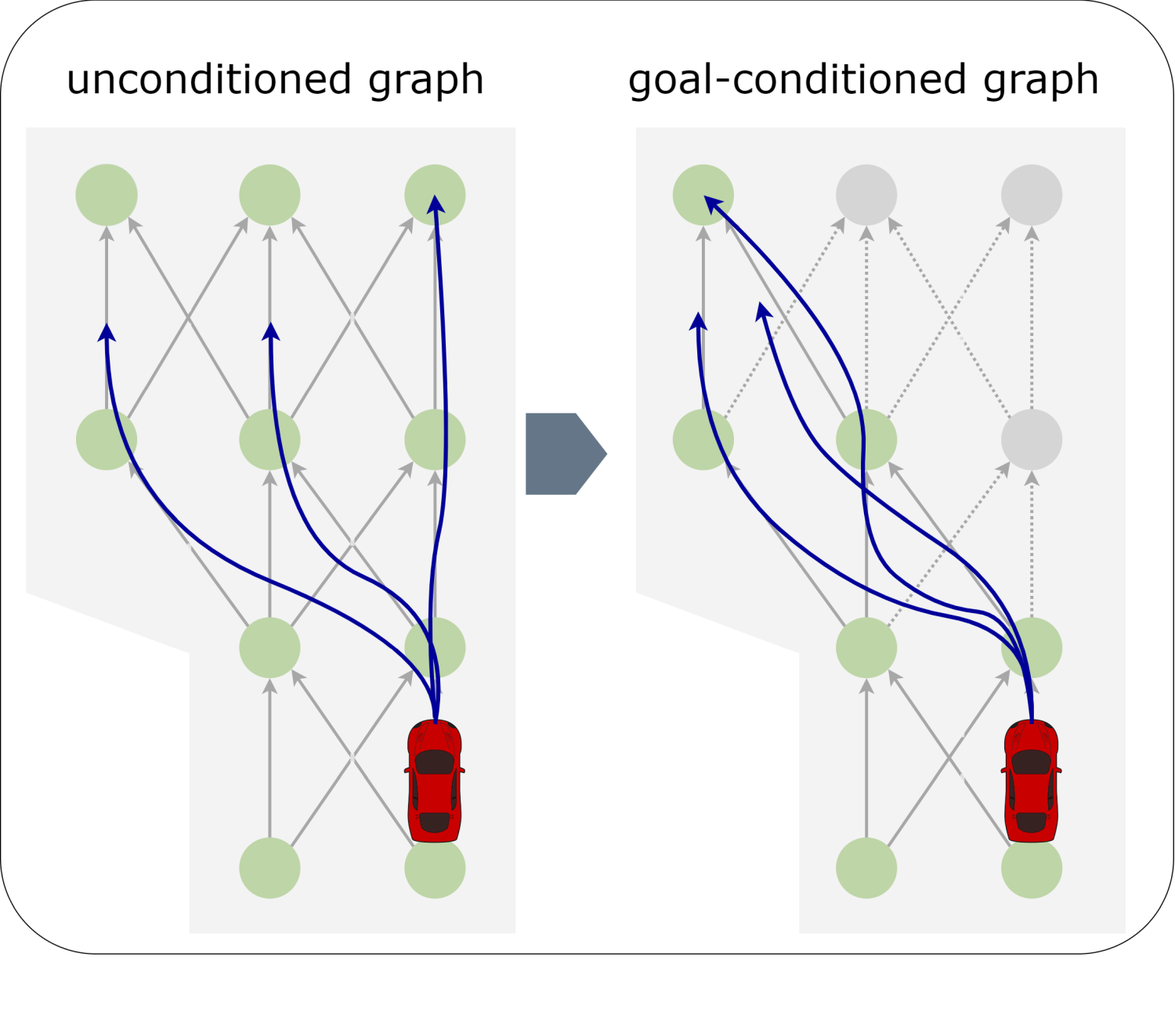}
\caption{Left: Candidate plans on an unconditioned graph cover all feasible behaviours, including those that do not follow the route. \newline
Right: Goal-conditioning is achieved by excluding trajectories that traverse off-route graph nodes (grey) from being considered. Hence, the model can use its entire multimodality budget to find trajectories that comply with a navigation goal, potentially yielding better plans.
}
\label{fig:first_page}
\end{figure}

In this work, our goal is to incorporate intention into a state-of-the-art predictor, converting it into an effective planner.
This work builds on the existing vehicle trajectory prediction model PGP~\cite{deo2022multimodal} and introduces a novel method for goal conditioning that precludes plans going off-route from being considered.
At the same time, the model retains its diversity at the motion level, such that it doesn't blindly follow a goal without reasoning about safety and comfort.
Fig.~\ref{fig:first_page} illustrates the degrees of freedom before and after goal conditioning.
Our main contribution is as follows: We introduce a novel method for goal-conditioning which allows to leverage advances in the field of prediction for the task of planning.
We leverage our method to convert the unconditioned PGP prediction model~\cite{deo2022multimodal} into a goal-conditioned planner.
We evaluate this planner in open-loop and closed-loop simulation on the \nuplan open-source dataset~\cite{caesar2021nuplan} and observe that this simple yet effective modification alone can significantly reduce the gap towards state-of-the-art planning performance.

The remainder of this paper is structured as follows: 
First, we discuss related work and recent advances in prediction and planning.
Subsequently, in Section~\ref{method} we introduce a novel approach for goal-conditioned planning which is evaluated in Section~\ref{experiments}.
Finally, Section~\ref{conclusions} draws conclusions and gives an outlook on promising future research directions.

\section{Related Work}
\label{related_work}

\subsection*{Prediction and Planning for Automated Driving}
The planning task aims to find a single trajectory that is optimal in terms of driving comfort, safety, efficiency, progress along the route, etc.
Besides rule-based approaches, which can suffer from poor generalization to new scenarios and need large engineering efforts, learning-based approaches have gained traction over recent years.
Pioneering work in the 1980s~\cite{pomerleau1988alvinn} demonstrated lane following on public roads with a shallow dense neural network using behaviour cloning.
A recent line of work leverages neural networks in optimization-based planning by learning cost functions from data~\cite{zeng2019end, wei2021perceive, sadat2020perceive, casas2021mp3}.
The trajectory prediction task is closely related to the task of planning, often using similar input and output representations.
A large body of literature exists, with new models showing ever-increasing performance on public benchmarks~\cite{lu2022kemp,varadarajan2022multipath++,liang2020learning,nayakanti2022wayformer,ye2022dcms,ye2021tpcn,shi2022mtr}.
Despite their similarities, prediction and planning are often tackled as separate tasks using different model architectures.
This makes it difficult for the field of planning to benefit from advances in the field of prediction.
In this work, we show how an existing state-of-the-art prediction model can be repurposed for the task of planning, directly carrying over advances from one task to the other.

\subsection*{From Grids to Graphs}
Representing the environment in the form of rasterized grids in combination with a convolutional neural network backbone has been a very popular approach for trajectory prediction~\cite{cui2019multimodal, gilles2021home, luo2018fast, chai2019multipath, phan2020covernet}.
In recent years a new generation of models emerged, that relies on graph-based representations for the map \cite{varadarajan2022multipath++,gilles2022gohome, liang2020learning,deo2022multimodal} and leverages graph neural networks or attention blocks to model interactions~\cite{gilles2021thomas, yuan2021agentformer, khandelwal2020if, casas2020spagnn, gao2020vectornet, knittel2022dipa, nayakanti2022wayformer}.
We refer the reader to~\cite{liu2021survey} for a comprehensive survey.
This innovation has not yet completely carried over to the planning task, with~\cite{scheel2022urban, wang2022ltp} being notable exceptions.
By repurposing a state-of-the-art prediction model for the task of planning, we are able to directly transfer recent advances including model architectures and input representations from prediction to planning.

\subsection*{Goal Conditioning}
A key challenge to repurposing a prediction model for planning is how to condition the diverse, multi-modal outputs on a given intention, \textit{i.e.}~the navigation goal.
The two predominant alternatives are to either switch between specialized submodels depending on a high-level command or add additional model inputs~\cite{codevilla2018end}.
The former is an effective way to reduce the imbalance in the training set and has been successfully applied in~\cite{casas2021mp3}.
However, as  the number of high-level commands and their definition have to be fixed in advance, this comes at the cost of flexibility.
Hence, the latter approach, \textit{i.e.} passing the intention as an additional input to the model, has been applied more broadly~\cite{chitta2021neat,Prakash2021CVPR,shao2022safety,chen2022learning}.
As already observed in~\cite{codevilla2018end} these models are only implicitly conditioned on the intention, with the risk that the route is not always followed.
Our approach circumvents this risk by conditioning directly on the intention level.
Another way to encourage goal-directed planning is the addition of a cost term that favours progress along the route~\cite{sadat2020perceive}.
The optimizer then has to find a balance between progress and other objectives such as safety and comfort, possibly sacrificing one for the other.

\subsection*{Goal-Conditioned Prediction as Planning}
Two-step prediction approaches separate the level of intended behaviour from the level of exact motion trajectory, instead of directly regressing the latter.
Various representations for the behaviour level have been proposed.
\cite{wang2022ltp} decodes trajectories conditioned on a target lane,
whereas~\cite{zhao2021tnt, gu2021densetnt} use the same paradigm but condition on concrete spatial locations or sparsely sampled trajectories~\cite{lu2022kemp}.
We build our work on~\cite{deo2022multimodal}, which conditions predictions on a coarse path represented by a traversal of the lane graph.
Since this path does not include any temporal information, such as the velocity profile or exact spatial locations, it represents a behaviour-level plan that can be followed with multiple motion trajectories.  
However, the approach from~\cite{deo2022multimodal} does not consider routing in the traversal selection.
We show that this separation of intention and motion level opens up an opportunity to define a new method of goal conditioning.
In particular, we demonstrate that by goal-conditioning of the traversals, the self-driving vehicle (SDV) can only consider route-compliant plans, while still being able and required to reason about safety in order to plan the exact trajectory.

\section{Method Description}
\label{method}

\begin{figure*}
\centering
\includegraphics[width=0.9\textwidth]{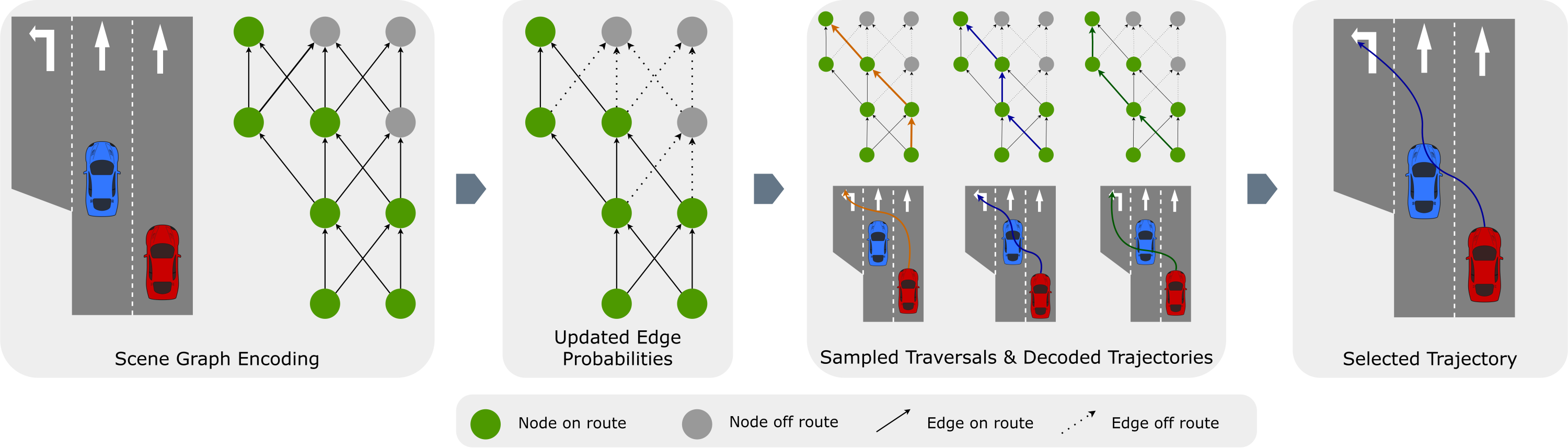}
\caption{Method overview: We manipulate the edge probabilities of a graph-based scene representation. Route-conditioned trajectories are then decoded from the manipulated graph and the most likely one selected as a plan for the SDV.}
\label{fig:model_overview}
\end{figure*}
\subsection{Graph-Based Prediction Model}
We give a brief recap of the unconditioned PGP prediction model~\cite{deo2022multimodal}, which we convert into a goal-directed planner by goal-conditioning at the level of intended behaviour. An overview of the PGP model architecture can be found in Fig.~\ref{app:pgp_model} in the appendix. 
We refer the reader to~\cite{deo2022multimodal} for a thorough description of the graph representation and details on the prediction model.
\subsubsection{Encoder}
The encoder builds a graph representation of the environment centred around a road graph $G$ consisting of nodes $V$ and directed edges $E$, i.e., $G(V,E)$.
The graph nodes $V$ represent the lane centrelines.
Therefore, lanes are divided into snippets of similar length and discretized into a polyline with a fixed maximum length.
Hence, the node features are a sequence of poses describing the centreline snippet.
Binary flags for stop lines and crosswalks are added to the poses.
Two types of directed edges $E=E_\text{succ} \cup E_\text{prox}$ model allowed transitions between the nodes $V$:
$E_\text{succ}$ for successor nodes along the lane in the direction of traffic flow and
$E_\text{prox}$ for proximal nodes on neighbouring lanes.

For each agent $i$ in the scene its history is represented as a trajectory
$s^i = {[x^i_t, y^i_t, v^i_t, a^i_t, \omega^i_t, I^i]}^0_{t=-t_h}$,
where $x^i_t, y^i_t$ are the position given in the SDV's local cartesian coordinate frame and $t$ is the timestep within the observation horizon $t_h$.
$v^i_t$, $a^i_t$, $\omega^i_t$ denote the agent's speed, acceleration, and yaw rate, respectively.
$ I^i$ is an indicator for agent class (vehicle or pedestrian).

Polylines, agents' histories, and SDV history are all encoded using gated recurrent units (GRU),
yielding encodings $h_{\text{node}}$, $h^i_{\text{agent}}$, and $h_{\text{SDV}}$.
Agent-to-node attention allows the lane nodes to accumulate traffic information,
before graph neural network (GNN) layers are applied to the road graph.
The full encoding is given by $h^v_{\text{agg}} \,\forall v \in V$.

\subsubsection{Graph Traversals}
The "aggregator" module determines transition probabilities for all edges and samples traversals through the graph.
Thus, for each node $u \in V$, the probabilities for all outgoing edges $(u,v)$ are obtained from $\text{score}(u,v)=\text{MLP}(h^u_{\text{agg}},h^v_{\text{agg}},h_{\text{SDV}})$ by \begin{equation}
\pi(u,v) = \softmax\left(\text{score}(u,v)) \,\vert\, (u,v) \in E \right)\,.
\end{equation}
Next, graph traversals are obtained by first assigning the SDV to the closest node and then iteratively sampling an outgoing edge and assigning the SDV to the next node along this edge.
In order to allow traversals of arbitrary length (up to a maximum number of $T$ nodes), terminal edges are added to each node.
At inference time, $K$ traversals
$\{v^{k}_n\}^T_{n=0}$, $k=1,..,K$
are sampled.

\subsubsection{Trajectory Decoder}
The traversal encoding is
\begin{equation}
h^k_{\text{traversal}} = \phi(h^v_{\text{agg}} \,\forall v \in \{v^k_n\}^T_{n=0})\,,
\end{equation}
where $\phi$ denotes concatenation. A latent variable model then decodes the predicted trajectory
\begin{equation}
o^k = \text{MLP}( \phi( h^k_{\text{traversal}}, h_{\text{SDV}}, z) )\,
\end{equation}
from a given traversal and a randomly sampled noise vector $z$.
In practice, the number of sampled traversals $K$ is large, and a fixed number of output trajectories is obtained after $k$-means clustering. Each trajectory is assigned a probability based on the respective cluster's rank.

\subsection{Route-Conditioned Traversals}
Our goal is to ensure goal conditioning of the trajectory output at the behaviour level.
To this end, we manipulate the edge probabilities obtained by the prediction model in a way that makes traversals that follow the intended route more likely.
An overview of our method is depicted in~Fig.~\ref{fig:model_overview}.
All nodes that lie on a path between the SDV's current position and the navigation goal are considered "on route", and the corresponding edges form the set $E_\text{route}$.

We propose two alternative methods for conditioning road-graph traversals.
First, we increase edge probabilities that stay on the route.
Second, we set probabilities of edges that deviate from the route to zero.
To enhance probabilities of edges that stay on the route, we add an additive bonus $\beta$ to the regressed score
\begin{equation}
\pi(u.v)=
\begin{cases}
    \softmax( \text{score}(u,v) ) + \beta& (u,v) \in E_\text{route} \\
    \softmax( \text{score}(u,v) ) & (u,v) \notin E_\text{route}
\end{cases}
\label{eq:soft_mask}
\end{equation}
The magnitude of the bonus $\beta$ is identical for all edges and is a learnable model parameter.
We call this setup \textit{soft-mask} goal conditioning.

Alternatively, we apply a \textit{hard mask} by setting the probabilities of edges that do not follow the route to zero.
This provides a guarantee that all traversals comply with the navigation goal.
The updated probabilities are given by
\begin{equation}
\pi(u.v)=
\begin{cases}
    \softmax( \text{score}(u,v) ) & (u,v) \in E_\text{route} \\
    0 & (u,v) \notin E_\text{route}
\end{cases}
\label{eq:hard_mask}
\end{equation}
We can apply the hard mask at inference time only (subsequently referred to as \textit{Goal-Conditioned PGP (GC-PGP)}), or already during training (termed \textit{hard mask}).
Applying the mask only at inference time is also beneficial, because the model still learns the entire multi-modal distribution, and no re-training is necessary.

\subsection{Trajectory Selection}
For a prediction model, it is a natural choice to output multiple trajectories per agent in order to account for the multi-modality induced by the unknown intention.
In contrast, for the planning task, the model has to commit to a single trajectory.
Similar to~\cite{pini2022safe} we apply a simple heuristic and select the trajectory with the highest probability.

Note that this choice does not necessarily correspond to the predominant road graph traversal (due to the clustering step).

\section{Experiments}
\label{experiments}
\subsection{Dataset and Evaluation Framework}
\label{experiments_dataset}
We employ the \nuplan dataset and framework~\cite{caesar2021nuplan} for model training and evaluation.
The dataset includes diverse scenarios, e.g.\ making turns, yielding to pedestrians, stopping at intersections, traversing intersections and driving at different speeds.
The full \nuplan train split consists of more than 30 million scenarios totalling over 1,300 hours of real-world driving.
For training, we use all of the available 70 scenario types at a maximum of 4,000 scenarios per type.
For some types, less than 4,000 scenarios are available, resulting in approx.\ 150,000 training scenarios.

The \nuplan entire test split consists of 4,539 scenarios.
We noticed annotation errors in some of these samples, which results in relevant lanes being labelled as off-route and vice versa.
Since this is particularly harmful in the context of target conditioning, we remove these samples from the test set.
By sampling a maximum of 30 scenarios for each available scenario type and removing the ones that are compromised by labelling errors, we obtain a test set consisting of 1,363 diverse driving scenarios.

In addition, we report results on a subset that focuses on scenarios where knowing the navigation goal is crucial.
We limit this reduced test set to the following four scenario types at 100 scenarios each: traversing intersection, starting unprotected non-cross turn, starting protected cross turn, and starting right turn. After removing the compromised samples, this intersection test split consists of 323 scenarios.

\subsection{Metrics}
\label{experiments_metrics}
Besides the aggregated driving score defined by the \nuplan framework, we also report average displacement error (ADE), final displacement error (FDE), and miss rate (MR) for the open-loop evaluations.
MR is defined by the final waypoint of the 8 second planning horizon deviating more than 16 metres from the ground truth.
For the closed-loop simulations, we evaluate progress, driveable area compliance, and collision avoidance.
Progress is given by the fraction of the travelled distance along the ground truth path.
Driveable area compliance describes the share of frames where the SDV's entire bounding box is on the driveable surface.
In order to evaluate collision avoidance, we report the fraction of successful scenarios in terms of at-fault collisions.
The SDV is considered to be at fault for a collision if it collides with a stationary agent or if the collision occurs at its front or sides.
We train and evaluate each model twice, and report the mean and standard deviation for all metrics.

\subsection{Baselines and Ablations}
\textit{Intelligent Driver Model baseline:}
IDM~\cite{treiber2000congested} is a heuristic model that keeps a safe distance from the vehicle ahead.
Its extension MOBIL~\cite{kesting2007general} also handles lane changes.
We use the implementation provided by the \nuplan framework~\cite{caesar2021nuplan}.

\textit{Urban Driver baseline:}
We compare our model to a state-of-the-art planning approach, which has been made available together with the \nuplan framework.
Urban Driver~\cite{scheel2022urban} uses PointNet layers to encode the SDV's motion as well as observed agents' history and map elements.
Subsequently, the information is fused using multi-head attention. Then a trajectory for the SDV is decoded from the aggregated encoding.
In contrast to our method, for this baseline, the route information is encoded into the map features.
The model is fairly large (2.2M parameters), so we adapt its complexity for improved comparability to PGP (150k parameters).
We reduce the encoding size from $256$ to $128$ and the number of subgraph layers from $3$ to $2$, resulting in 520k parameters (\textit{UD-520k}).
Similarly, the \textit{UD-150k} model uses an encoding size of $64$, but retains all $3$ subgraph layers, totalling 150k parameters.

\textit{Filter on route baseline:}
The multi-modal output of PGP is post-conditioned on the route:
To this end, we only keep trajectories whose endpoint is within 5 metres of the nearest lane centre on the route;
among those, we pick the trajectory with the largest probability.
If no trajectory is within the distance threshold, we pick the one closest to the route.

\textit{Node features:}
An alternative to explicitly conditioning a model on the navigation goal is to extend the model's input features~\cite{codevilla2018end}.
Hence, we add a binary value indicating whether the node is part of the route to each node in the lane graph.

\textit{Soft mask:}
The transition probabilities of edges that keep the traversal on route receive a (learnable) bonus,
see eq.~\eqref{eq:soft_mask}.

\textit{Hard mask:}
As opposed to our GC-PGP model, edges that leave the route are removed, also at train time (see eq.~\eqref{eq:hard_mask}).
This way both the aggregator and trajectory decoder models are restricted to only goal-reaching traversals.

\subsection{Open-Loop Simulation}
\label{experiments_ol}

\begin{figure*}
\centering
\includegraphics[width=\textwidth]{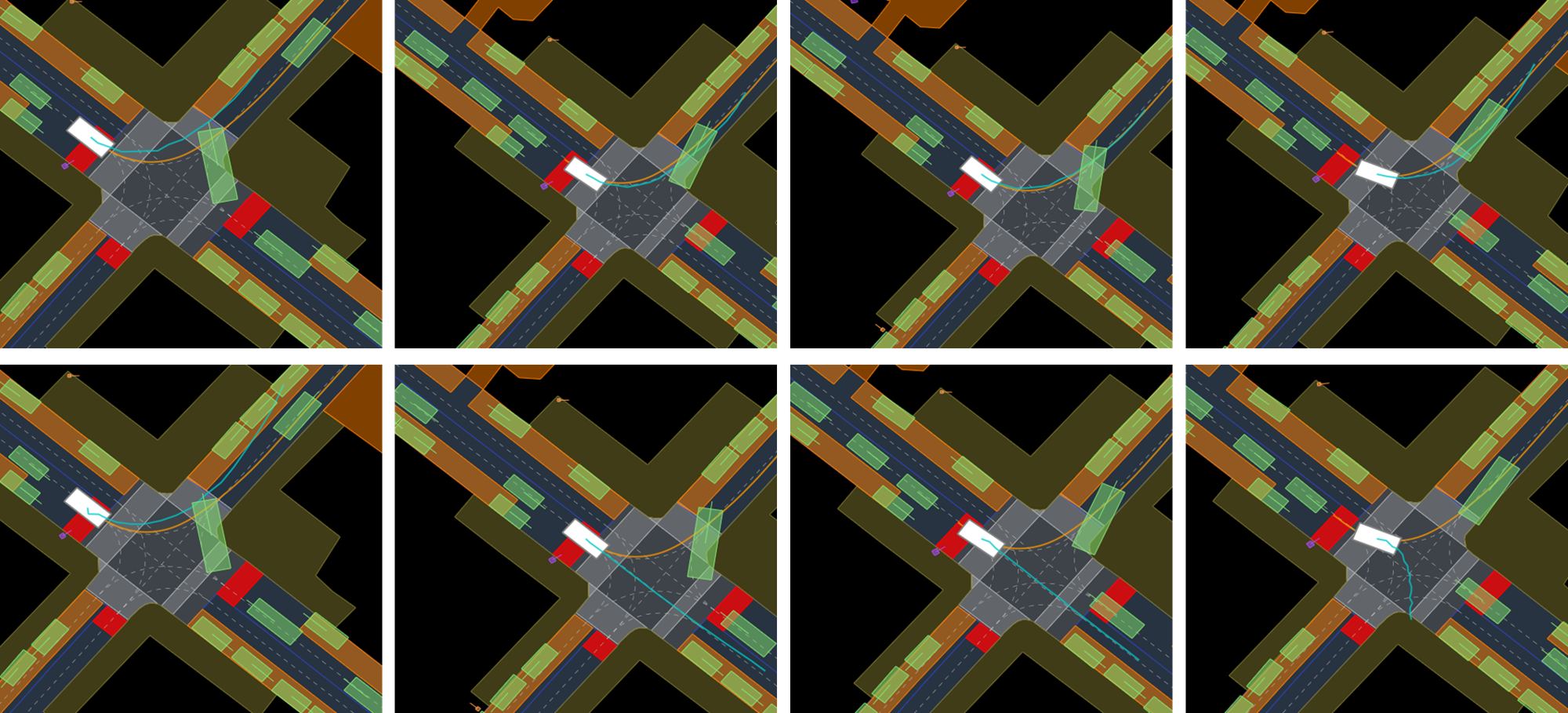}
\caption{Qualitative results: Left to right depicts the first 4 seconds of the SDV making a left turn. Top: GC-PGP, Bottom: \textit{node features} ablation. The expert trajectory and model prediction are depicted in orange and blue respectively. The orange surface indicates parking areas. Stop lines are shown in red.
In the bottom row toggling between turning and going straight manoeuvres can be observed, whereas our method (top) results in plans that are compliant with the route at all times.
(Best viewed in colour.)}
\label{fig:qualitative_results}
\end{figure*}

\begin{table}[ht]
\centering
\caption{Open-loop evaluation on the entire test split}
\label{tab:openloop_overall}
\adjustbox{max width=\the\columnwidth}{
\begin{tabular}{ lllll }
\toprule
\textbf{Model}            & \textbf{Score}  & \textbf{ADE [m]}& \textbf{FDE [m]}& \textbf{MR}\\
\midrule
IDM/MOBIL                 & $0.33$          & $7.98$          & $12.8$          & $0.42$\\
Urban Driver              & $0.84 \pm 0.01$ & $1.42 \pm 0.00$ & $2.99 \pm 0.07$ & $0.06 \pm 0.00$\\
UD-520k                   & $0.83 \pm 0.00$ & $1.49 \pm 0.03$ & $3.18 \pm 0.08$ & $0.06 \pm 0.02$\\
UD-150k                   & $0.80 \pm 0.01$ & $1.62 \pm 0.02$ & $3.39 \pm 0.00$ & $0.08 \pm 0.00$\\
PGP                       & $0.69 \pm 0.02$ & $1.82 \pm 0.06$ & $4.21 \pm 0.14$ & $0.14 \pm 0.00$\\
GC-PGP (ours)             & $0.76 \pm 0.01$ & $1.59 \pm 0.03$ & $3.65 \pm 0.09$ & $0.12 \pm 0.01$\\
\midrule
Filter on route           & $0.73 \pm 0.01$ & $1.66 \pm 0.01$ & $3.84 \pm 0.03$ & $0.13 \pm 0.00$\\
Node features             & $0.73 \pm 0.01$ & $1.70 \pm 0.02$ & $3.91 \pm 0.02$ & $0.13 \pm 0.00$\\
Soft mask                 & $0.71 \pm 0.00$ & $1.75 \pm 0.00$ & $4.02 \pm 0.05$ & $0.14 \pm 0.01$\\
Hard mask                 & $0.62 \pm 0.04$ & $2.23 \pm 0.27$ & $4.82 \pm 0.99$ & $0.17 \pm 0.05$\\
\bottomrule
\end{tabular}
}
\end{table}

\begin{table}[ht]
\centering
\caption{Open-loop evaluation on intersection scenarios}
\label{tab:openloop_intersections}
\adjustbox{max width=\the\columnwidth}{
\begin{tabular}{ lllll }
\toprule
\textbf{Model}            & \textbf{Score}  & \textbf{ADE [m]}& \textbf{FDE [m]}& \textbf{MR}\\
\midrule
IDM/MOBIL                 & $0.19$          & $4.89$          & $5.81$          & $0.20$\\
Urban Driver              & $0.83 \pm 0.00$ & $1.66 \pm 0.02$ & $3.36 \pm 0.03$ & $0.05 \pm 0.00$\\
UD-520k                   & $0.81 \pm 0.00$ & $1.80 \pm 0.04$ & $3.73 \pm 0.09$ & $0.07 \pm 0.00$\\
UD-150k                   & $0.77 \pm 0.00$ & $1.96 \pm 0.04$ & $4.03 \pm 0.08$ & $0.08 \pm 0.00$\\
PGP                       & $0.56 \pm 0.03$ & $2.48 \pm 0.08$ & $5.67 \pm 0.20$ & $0.19 \pm 0.01$\\
GC-PGP (ours)             & $0.74 \pm 0.01$ & $1.80 \pm 0.04$ & $4.03 \pm 0.11$ & $0.10 \pm 0.00$\\
\midrule
Filter on route           & $0.68 \pm 0.01$ & $1.98 \pm 0.01$ & $4.50 \pm 0.04$ & $0.14 \pm 0.00$\\
Node features             & $0.67 \pm 0.02$ & $2.15 \pm 0.07$ & $4.81 \pm 0.12$ & $0.14 \pm 0.01$\\
Soft mask                 & $0.60 \pm 0.00$ & $2.41 \pm 0.00$ & $5.43 \pm 0.05$ & $0.19 \pm 0.01$\\
Hard mask                 & $0.67 \pm 0.04$ & $2.20 \pm 0.15$ & $4.92 \pm 0.29$ & $0.16 \pm 0.03$\\
\bottomrule
\end{tabular}}
\end{table}

We report in Tab.~\ref{tab:openloop_overall} open-loop imitation performance on the full, diverse test split.
This dataset is dominated by scenarios where route information is not required, such as being stationary at traffic lights or simple lane following.
We observe that all goal-conditioning variants still outperform the (undirected) PGP baseline in all metrics, closely approaching the strong Urban Driver baseline.
Interestingly enough the IDM/MOBIL model is a poor contestant w.r.t.\ imitation metrics: its behaviour may have desirable properties, but it is not particularly human-like on \nuplan scenarios.

To demonstrate the effectiveness of our approach we also evaluate on a reduced test set as described in Sec.~\ref{experiments_dataset}.
This dataset is focused on non-trivial intersection scenarios where the navigation goal is important to the driving behaviour.
Results are reported in Tab.~\ref{tab:openloop_intersections} and confirm our previous observations.
Our goal-conditioned model (GC-PGP) clearly outperforms the PGP prediction baseline and catches up to the Urban Driver planner baseline in overall score and displacement error.

Looking at alternative goal-conditioning methods we observe that the \textit{node features} and \textit{soft mask} models only show mild improvements over PGP.
We speculate that they do not always strictly adhere to the route information, a claim that is supported by qualitative results in Fig.~\ref{fig:qualitative_results}.
In contrast, the masking in our GC-PGP model prevents taking plans off the route into consideration.
However, we note that applying this hard mask also at train time to focus on goal-directed traversals only compromises performance.
This model is much less robust against route labelling errors as described in Sec.~\ref{experiments} and we hypothesize about a connection between the resulting faulty mask and observed instabilities during training.

\subsection{Temporal Plan Stability}
\begin{figure*}[h!]
\centering
\includegraphics[width=\textwidth]{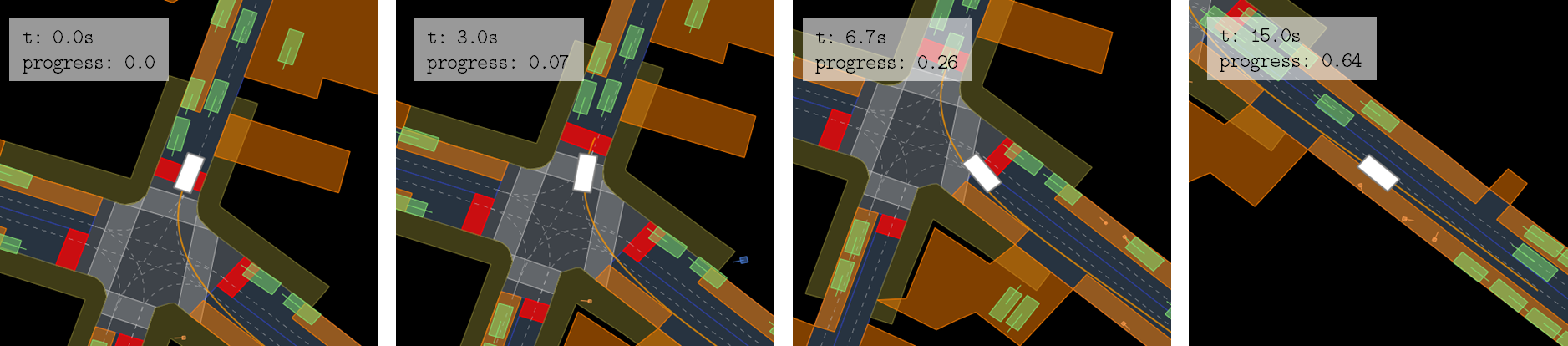}
\caption{Making a turn in closed-loop simulation. The SDV enters the intersection after 3 seconds having travelled $7$\% of the expert's total travelled distance and leaves the intersection after 6.7 seconds with $26$\% progress. After 15 seconds of simulation, it covers $64$\% of the expert's path. (Best viewed in colour.)}
\label{fig:cl_qualitative_results}
\end{figure*}
\begin{table}[ht]
\centering
\caption{Temporal plan instability on intersection scenarios}
\label{tab:toggling}
\begin{tabularx}{\columnwidth}{Xl}
\toprule
\textbf{Model}            & \textbf{TPI [m]} \\
\midrule
IDM/MOBIL                 & $0.58$\\
Urban Driver              & $1.51 \pm 0.00$\\
UD-520k                   & $1.78 \pm 0.02$\\
UD-150k                   & $1.84 \pm 0.14$\\
PGP                       & $3.65 \pm 0.15$\\
GC-PGP (ours)             & $2.80 \pm 0.00$\\
\bottomrule
\end{tabularx}
\end{table}

The deviation between planned trajectories for consecutive timesteps is a measure of the robustness of the planner.
A low value means that the plans are consistent over time, indicating robustness and resulting in a comfortable driving experience.
We define temporal plan instability (TPI)
\begin{align}
\mathrm{tpi}(\tau) = \| \left(x,y\right)_T^{@ \tau} - \left(x,y\right)_{T-1}^{@ \tau+1} \|_2
\end{align}
the distance between (adjusted) trajectory end-points for consecutive timesteps ($@\tau$, $@\tau+1$)
and report results in Tab.~\ref{tab:toggling}.

The IDM/MOBIL baseline produces the most time-consistent plans and can serve as a lower limit, which is unsurprising given its underlying simple heuristics.
PGP is the least time-consistent model.
As a prediction model, it has been designed to produce diverse trajectories, both w.r.t.\ the route taken and different velocity profiles.
Toggling between these modes naturally leads to large plan instability.
Conditioning on a navigation goal (GC-PGP) removes the main source of multi-modality, but diversity w.r.t.\ velocity remains as a cause for instability.
GC-PGP thus yields more stable plans compared to PGP but retains a higher degree of toggling than Urban Driver, an inherently unimodal planning model.
This indicates that the na\"ive way we select a trajectory among the multi-modal proposals may not be ideal.
Consequently, planning performance could be further improved by applying a cost function that balances different aspects of driving, such as safety, comfort, progress, etc., as in~\cite{sadat2020perceive, casas2021mp3}.

\subsection{Closed-Loop Simulation}
\begin{table}[ht]
\centering
\caption{Closed-loop evaluation on intersection scenarios}
\label{tab:closed_loop}
\adjustbox{max width=\the\columnwidth}{
\begin{tabular}{ lllll }
\toprule
\textbf{Model}            & \textbf{Score}  & \textbf{Progress}& \textbf{Driv. Area}& \textbf{Col.-Free}\\
\midrule
IDM/MOBIL                 & $0.75$          & $0.81$           & $0.96$             & $0.90$\\
Urban Driver              & $0.58 \pm 0.06$ & $0.76 \pm 0.11$  & $0.89 \pm 0.02$    & $0.85 \pm 0.01$\\
UD-520k                   & $0.50 \pm 0.08$ & $0.71 \pm 0.06$  & $0.90 \pm 0.02$    & $0.84 \pm 0.02$\\
UD-150k                   & $0.45 \pm 0.02$ & $0.74 \pm 0.08$  & $0.81 \pm 0.01$    & $0.79 \pm 0.03$\\
PGP                       & $0.37 \pm 0.05$ & $0.36 \pm 0.07$  & $0.88 \pm 0.04$    & $0.89 \pm 0.03$\\
GC-PGP (ours)             & $0.46 \pm 0.03$ & $0.47 \pm 0.08$  & $0.88 \pm 0.01$    & $0.88 \pm 0.04$\\
\midrule
Filter on route           & $0.43 \pm 0.02$ & $0.43 \pm 0.09$  & $0.87 \pm 0.04$    & $0.87 \pm 0.01$\\
Node features             & $0.37 \pm 0.00$ & $0.37 \pm 0.03$  & $0.86 \pm 0.05$    & $0.87 \pm 0.03$\\
Soft mask                 & $0.29 \pm 0.02$ & $0.30 \pm 0.06$  & $0.91 \pm 0.00$    & $0.92 \pm 0.02$\\
Hard mask                 & $0.43 \pm 0.04$ & $0.43 \pm 0.06$  & $0.84 \pm 0.01$    & $0.84 \pm 0.01$\\
\bottomrule
\end{tabular}}
\end{table}

We also evaluate all models in the log-playback closed-loop simulation environment provided by the \nuplan framework.
The same reduced test set as for the open-loop evaluation is used (cf.\ Sec.~\ref{experiments_dataset}).
Results are presented in Tab.~\ref{tab:closed_loop}.
It is striking that the hand-crafted IDM/MOBIL model achieves the best overall performance by a large margin,
but is still not perfectly safe in terms of drivable area compliance and caused collisions.
This completes the picture we gained from the open-loop evaluation (Sec.~\ref{experiments_ol}): IDM/MOBIL is a strong "classical" baseline, just not in terms of expert imitation.

Both PGP and GC-PGP are on par with Urban Driver regarding drivable area compliance, and even outperform w.r.t.\ caused collisions.
Although GC-PGP improves upon (the undirected) PGP in terms of progress along the route, with a value of $0.47$ it still falls behind Urban Driver.
See Fig.~\ref{fig:cl_qualitative_results} to put this value into perspective:
At some point well behind the crossing, GC-PGP fails to progress along the lane fast enough.
We conclude that an average progress of only $0.47$ is on average sufficient to master the crossing, but indicates shortcomings in the subsequent lane following task, which exceeds the scope of our method.

For the alternative goal-conditioning approaches (lower part of Tab.~\ref{tab:closed_loop}), we do not observe significant deviations from their PGP and GC-PGP counterparts w.r.t.\ drivable area compliance and caused collisions.
Instead, we do find that they all achieve less progress than GC-PGP, the \textit{soft mask} variant being even worse than the PGP baseline.
These results confirm GC-PGP as the most performant variant to condition PGP on a navigation goal.

\section{Conclusions}
\label{conclusions}
In this work, we presented a novel way to achieve goal conditioning on a graph-based map representation that precludes behaviour-level plans going off-route from being considered.
We demonstrated the effectiveness of our approach in a comprehensive benchmark, by repurposing a state-of-the-art prediction model for the planning task and comparing it against rule-based and data-based models, thereby highlighting that goal-conditioned prediction models can serve as strong baselines in planning benchmarks and as effective starting points for novel planning methods.
While our method reduces the multimodality at the behaviour level, the planner still needs to consistently select from multi-modal motion trajectory candidates.
We will address this uni-modality problem in future work.

\bibliographystyle{IEEEtran}
\bibliography{IEEEabrv,bibliography}

\clearpage
\onecolumn
\appendix
\subsection{PGP Model}
\begin{figure*}[h!]
    \centering
    \includegraphics[width=0.9\textwidth]{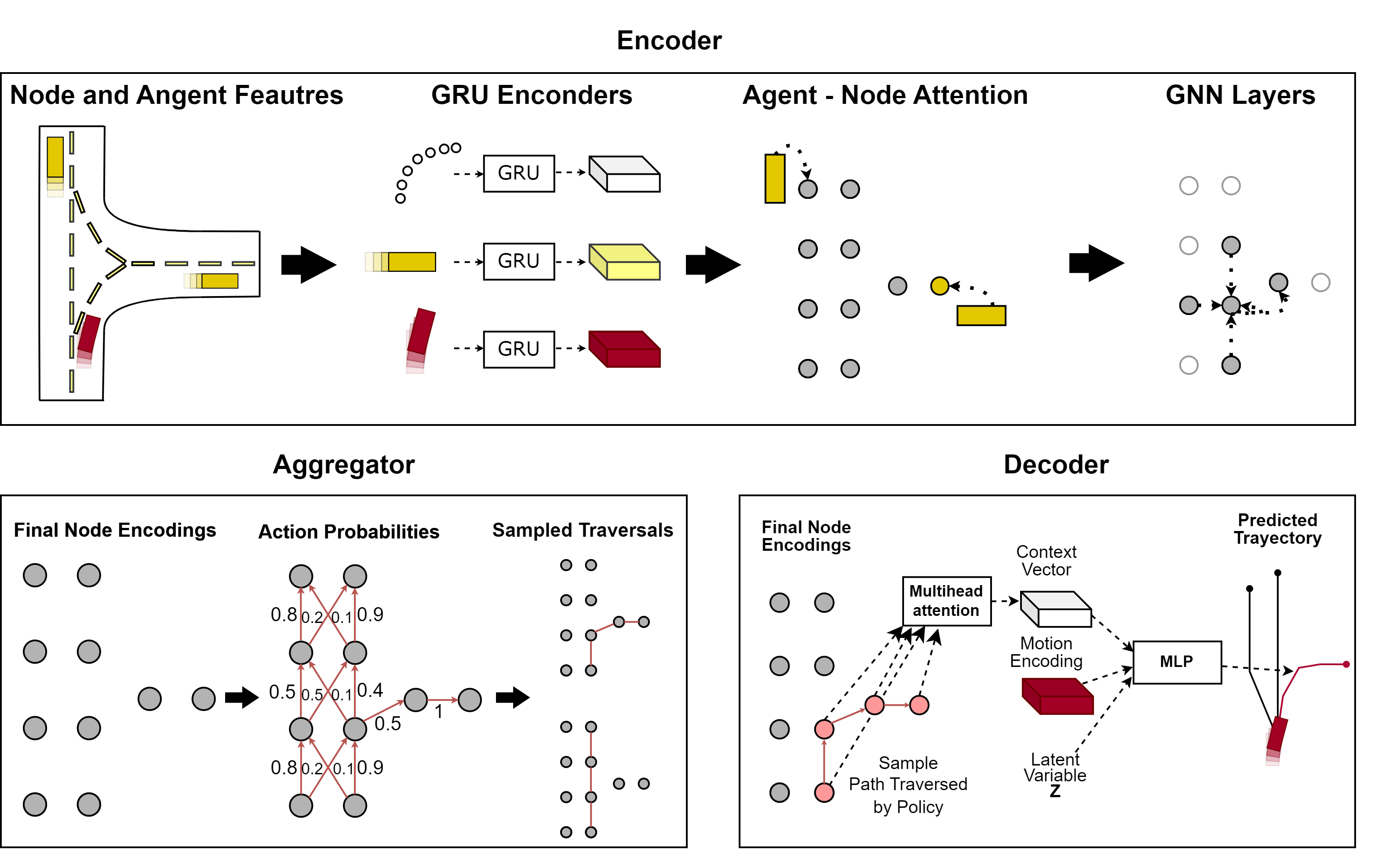}
    \caption{\textbf{Overview of the PGP Model.} Adapted from~\cite{deo2022multimodal}.
    The model consists of three modules. First, the encoder takes the history of surrounding agents as well as the SDV motion and map information as input and encodes with Gated Recurrent Units (GRU). Agent-Node Attention and GNN layers generate a road graph containing real-time traffic information.
Subsequently, the policy header predicts probabilities for the outgoing edges of each graph node. Sampling these probabilities yields traversals, that describe a likely future behaviour.
Finally, the latent-variable decoder regresses the exact motion trajectory based on the traversals.
    }
    \label{app:pgp_model}
\end{figure*}
\end{document}